# An effective Genetic Programming Hyper-Heuristic for Uncertain Agile Satellite Scheduling

1st Yuning Chen
College of Systems Engineering
National University of Defense Technology
Changsha, China
yuning_chen@nudt.edu.cn

2nd Junhua Xue*
College of Systems Engineering
National University of Defense Technology
Changsha, China
1325438117@qq.com

3rd Wangqi Gu
College of Systems Engineering
National University of Defense Technology
Changsha, China
3042794341@qq.com

4th Mingyan Shao
College of Systems Engineering
National University of Defense Technology
Changsha, China
313916696@qq.com

*Abstract*—This paper investigates a novel problem, namely the Uncertain Agile Earth Observation Satellite Scheduling Problem (UAEOSSP). Unlike the static AEOSSP, it takes into account a range of uncertain factors (e.g., task profit, resource consumption, and task visibility) in order to reflect the reality that the actual information is inherently unknown beforehand. An effective Genetic Programming Hyper-Heuristic (GPHH) is designed to automate the generation of scheduling policies. The evolved scheduling policies can be utilized to adjust plans in real time and perform exceptionally well. Experimental results demonstrate that evolved scheduling policies significantly outperform both well-designed Look-Ahead Heuristics (LAHs) and Manually Designed Heuristics (MDHs). Specifically, the policies generated by GPHH achieve an average improvement of 5.03% compared to LAHs and 8.14% compared to MDHs.

*Keywords—Agile earth observation satellite scheduling problem, uncertain, Genetic Programming Hyper-Heuristic*

## I. INTRODUCTION

Earth Observation Satellite (EOS) is a space platform that can be configured to capture images of surface targets according to varying requirements [1]. EOSs can respond well to the demand for rich image data in areas such as disaster relief, environmental monitoring and agriculture, making them highly valuable in practice. However, EOSs remain relatively scarce as high-value strategic resources. The diverse and complex application scenarios impose higher demands on EOS scheduling. Non-agile EOSs possess only roll axis freedom during in-orbit operation. In contrast, Agile EOSs (AEOS) have three degrees of freedom (roll, pitch and yaw). AEOSs offer greater flexibility as well as the ability to adjust their posture rapidly within a short period of time. Therefore, they have a longer time window in which to observe each task, making observation more likely [2]. Unlike the EOS Scheduling Problem (EOSSP) which can be viewed as a selection problem, the AEOSSP requires observation tasks to be selected, as well as the observation time window for each task to be determined.

Traditional imaging satellite scheduling relies on ground control, resulting in bottlenecks such as lengthy scheduling cycles, poor timeliness, and weak resistance to interference. The rapid development of aerospace technology has put forward higher demands on the management and dispatching of satellites. In response, intelligence is now a key research direction in the field of satellite scheduling. Due to the significant improvements in onboard computing capabilities, modern satellites are transforming from passive actuators to intelligent entities. This means that scheduling will operate within an autonomous control loop of "perception-decision-execution". AEOS must face numerous uncertainties during task execution [3], [4]. Accordingly, compared with conventional scheduling methods, intelligent approaches offer greater practicality. The task profit may be various at different moments. The visibility of task changes due to meteorological factors [5]. The resource consumption of the task cannot be known in advance before it is observed. Under conditions where various uncertainties are considered, pre-planned solutions may deteriorate or even become unfeasible. Therefore, this paper focuses on seeking a highly efficient algorithm to solve an uncertain AEOSSP (UAEOSSP) and produce a solution with highest expected profit.

AEOS can utilize scheduling policy to carry out real-time decision making and realize autonomous task scheduling. Numerous studies on manually designed scheduling policies have demonstrated promising performance [6]. Meanwhile, the rapid development of artificial intelligence has introduced a novel paradigm for generating scheduling policies. Scholars have successfully trained neural networks based on reinforcement learning methods and deployed them as scheduling policies to enhance dynamic onboard scheduling. Inspired by recurrent neural networks and the attention mechanism, Chen et al. designed a reinforcement learning-based end-to-end framework model [7].

Although manually designed and neural network-based machine learning offer advantages in generating scheduling policies, they also have significant shortcomings. Firstly, formulating a manual scheduling policy is very time-consuming and requires expertise for implementation. Secondly, the robustness of manually designed policies across various scenarios is difficult to guarantee. In addition, neuro networks are "black box" models, which makes it difficult to analyze their decision-making logic. Finally, the training of neural network models relies very much on the training data and methods used. To address the aforementioned issues, Genetic Programming Hyper-Heuristic (GPHH) will be employed to automatically generate effective scheduling policies for any given UAEOSSP scenario.

*Corresponding author

GPHH is a machine learning method that combines functions and terminal elements intelligently through data-driven approaches, evolving formal policy functions. It has the capacity to automatically learn complex relationships among problem features while exhibiting dynamic adaptability [8], [9]. Individuals in the GP population are scheduling policies, typically represented as executable programs or expression trees. Previous studies have used GPHH to generate policies to solve varieties of combinatorial optimization problems and they reported good results [10]. Ho and Tay used GPHH to address the Flexible Job Shop scheduling problem [11]. Dimopoulos and Zalzala applied GPHH to solve the Uncertainty Considered Single Machine scheduling problem [12]. However, few studies have applied GPHH to satellite scheduling [13] and existing work has largely overlooked the incorporation of uncertain factors.

To advance the practical application of autonomous scheduling research for AEOSs, this paper proposes and constructs a UAEOSSP model capable of taking multiple uncertainties into account. Furthermore, a method for automatically generating robust scheduling policies will be designed to address the UAEOSSP. The main contributions of this research are as follows:

1) Build a UAEOSSP model that takes into account uncertainties in task profit, resource consumption and task visibility.

2) Propose a GPHH method that can automate the training of effective scheduling policies that are embedded in a dynamic scheduling algorithm to produce high quality solutions.

3) Conduct experiments to validate the proposed model and the effectiveness of GPHH.

## II. UNCERTAIN AGILE EARTH OBSERVATION SATELLITE SCHEDULING PROBLEM

### A. Problem Description and Assumptions

In the UAEOSSP, we are given a set of candidate tasks and its corresponding visible time windows, where task profit, resource consumption, and task visibility are uncertain. The purpose is to select a subset of tasks and schedule their observation times to maximize the expected total profit.

The input for UAEOSSP is a set of problem instances, which represent AEOS scheduling scenarios under different environmental conditions. In these instances, the memory consumption of a task remains unknown until it is observed. Consequently, during the scheduling process, the observed memory consumption may exceed the remaining memory on the satellite, resulting in "imaging failure". Since decisions are irreversible, "imaging failure" will lead to the loss of the task profit and exhaustion of the memory on the satellite.

The simplifications and assumptions given for this research problem are as follows:

- Satellite resources and the visible time window for tasks are both known.
- Only point tasks are considered [14].
- The AEOS adopts the uniform ground speed imaging mode, and the memory consumption is positively correlated with the imaging time.
- Cloud cover affects the imaging quality, which in turn influences the task profit. Therefore, the effect of partial cloud cover is not considered.

### B. Variables

The variables of the problem model are shown in the TABLE I. The decision variables for the UAEOSSP are $\bar{x}_i(env)$, $\overline{y_{ij}}(env)$, $\overline{os_i}(env)$ and $\overline{oe_i}(env)$.

TABLE I. THE PROBLEM VARIABLES OF UAEOSSP

| Notation | Description |
|---|---|
| $ws_i$ | The start time of visible time window for task $i$. |
| $we_i$ | The end time of visible time window for task $i$. |
| $att_{t,i}$ | The attitude information of the satellite observing the task $i$ at time $t$, including pitch, roll and yaw angle |
| $p_i$ | Profit of task $i$. |
| $du_i$ | The time required for task $i$ to be completely imaged. |
| $cr$ | Expected memory write code rate. |
| $MMC$ | Maximum memory capacity. |
| $NT$ | Number of tasks. |
| $E$ | A collection of scheduling scenarios across multiple environments. |
| $env$ | A scheduling scenario in a given environment. $env \in E$. |
| $\bar{p}_i(env)$ | The actual profit of task $i$ in $env$, which is a predictable variable. |
| $\overline{cr_i}(env)$ | The average write speed for imaging task $i$ in $env$, which is an unpredictable variable. |
| $\overline{vis_i}(env)$ | The visibility of task $i$ in $env$, where values 0 and 1 denote invisible and visible, respectively. It is a predictable variable. |
| $\bar{x}_i(env)$ | Whether to observe task $i$ in $env$. It is defined by Eq. (1). |
| $\overline{y_{ij}}(env)$ | Whether task $j$ is the next task to be observed after task $i$ in $env$. It is defined by Eq. (2). |
| $\overline{os_i}(env)$ | The time when the satellite starts to observe task $i$ in $env$. |
| $\overline{oe_i}(env)$ | The time when the satellite ends to observe task $i$ in $env$. |

$$\bar{x}_i(env) = \begin{cases} 1, & \text{if task } i \text{ is selected} \\ 0, & \text{otherwise} \end{cases} \quad (1)$$

$$\overline{y_{ij}}(env) = \begin{cases} 1, & \text{if task } j \text{ is selected} \\ & \text{immediately after task } i \\ 0, & \text{otherwise} \end{cases} \quad (2)$$

### C. Objective Function

The objective function (3) is to maximize the expected total profit over all environments.

$$\max \frac{\sum_{i=1}^{NT} \bar{x}_i(env) \cdot \bar{p}_i(env)}{|E|} \quad (3)$$

### D. Constraints

The actual amount of memory consumed must not exceed the maximum memory capacity.

$$\sum_{i=1}^{NT} (\bar{x}_i(env) \cdot \overline{cr_i}(env) \cdot du_i) \leq MMC \quad (4)$$

A task cannot be observed if it is invisible.

$$\bar{x}_i(env) = 0, \forall \overline{vis_i}(env) \quad (5)$$

If task $i$ is observed, the observation time window must be within its visible time window.

$$ws_i \leq \overline{os_i}(env) < \overline{oe_i}(env) \leq we_i, \forall \bar{x}_i(env) = 1 \quad (6)$$

If task $i$ is observed, the observation duration equals the time required for complete imaging.

$$\overline{oe_i}(env) - \overline{os_i}(env) = du_i, \forall \overline{x_i}(env) = 1 \quad (7)$$

The attitude transition time between two adjacent observations cannot be greater than the interval between observation time windows. The transition time between $att_{t1,i1}$ and $att_{t2,i2}$ is represented as $Trans(att_{t1,i1}, att_{t2,i2})$. $\Delta g$ is the transition angle between the two attitudes, and its calculation is as Eq. (9). $\gamma, \eta$ and $\vartheta$ respectively represent the pitch angle, roll angle, and yaw angle of satellite. The transition time can typically be presented by a piecewise linear function, as shown in Eq. (10) [15], [16].

$$\overline{oe_i}(env) + Trans\left(att_{oe_i,i}, att_{os_j,j}\right) \leq \overline{os_j}(env),$$
$$\forall \overline{y_{ij}}(env) = 1 \quad (8)$$

$$\Delta g = |\gamma(att_{t1,i1}) - \gamma(att_{t2,i2})| +$$
$$|\eta(att_{t1,i1}) - \eta(att_{t2,i2})| + |\vartheta(att_{t1,i1}) - \vartheta(att_{t2,i2})| \quad (9)$$

$$Trans(att_{t1,i1}, att_{t2,i2}) = \begin{cases} a_1 + \frac{\Delta g}{v_1}, & \Delta g \leq \theta_{11} \\ a_2 + \frac{\Delta g}{v_2}, \theta_{20} \leq \Delta g \leq \theta_{21} \\ \ldots \\ a_n + \frac{\Delta g}{v_n}, \theta_{n0} \leq \Delta g \leq \theta_{n1} \end{cases} \quad (10)$$

In Eq. (10): $a$ is a constant; $v$ is the attitude angle transition velocity of satellite; $\theta$ is the attitude transition angle.

There is no sequential relationship between any task and itself.

$$\overline{y_{ii}}(env) = 0 \quad (11)$$

In order to prevent the first and last tasks in the imaging scheme sequence from being inexpressible in $y$, virtual tasks are introduced. The value of virtual tasks at $x$ is constrained.

$$\overline{x_0}(env) = \overline{x_{NT+1}}(env) = 1 \quad (12)$$

For any task, there is only one successor and one predecessor task if it is imaged.

$$\overline{x_i}(env) = \sum_{j=1}^{NT} \overline{y_{ij}}(env) \quad (13)$$

$$\overline{x_j}(env) = \sum_{i=1}^{NT} \overline{y_{ij}}(env) \quad (14)$$

If the task $j$ is observed after the task $i$, both tasks need to be observed.

$$\overline{x_i}(env) = 1 \wedge \overline{x_j}(env) = 1, \forall \overline{y_{ij}}(env) = 1 \quad (15)$$

The DFJ constraint proposed by Dantzig et al. in 1954 [17] effectively avoids sub-loops on the path.

$$\sum_{i \in S} \sum_{j \in S} \overline{y_{ij}}(env) \leq |S| - 1, \forall S \subseteq \{1, 2, \ldots, NT\}, S \neq \emptyset \quad (16)$$

The domain constraints of decision variables.

$$\overline{x_i}(env) \in \{0,1\}, \overline{y_{ij}}(env) \in \{0,1\},$$
$$\overline{os_i}(env) \in R^+, \overline{oe_i}(env) \in R^+ \quad (17)$$

## III. GENETIC PROGRAMMING HYPER-HEURISTIC FOR UAEOSSP

### A. The overall framework of GPHH

Similar to other methods based on GPHH for solving scheduling problems, an evolution procedure is used to automatically generate scheduling policies, which are then embedded into a scheduling algorithm to evaluate the quality of the policies. The GPHH includes training stage and testing stage. In the training stage, the scheduling policies are generated by exploring and exploiting the policy space. In the testing stage, the actual performances of GP individuals are evaluated. The flowchart of GPHH is given in Fig.1, as well as an introduction to the process as follows:

Step 1: Initialize and generate a GP population containing $N$ individuals (corresponding to $N$ scheduling policies). The population is initialized using Half and Half [18].

Step 2: Evaluate policy performance based on a mini-batch, which comprises multiple training instances. A dynamic scheduling algorithm is designed to evaluate policies.

Step 3: Use genetic operators to generate offspring.

Step 4: Evaluate policy performance based on the test set.

Step 5: If the termination condition is satisfied, return the optimal scheduling policy found during the search process. Otherwise proceed to Step 2.

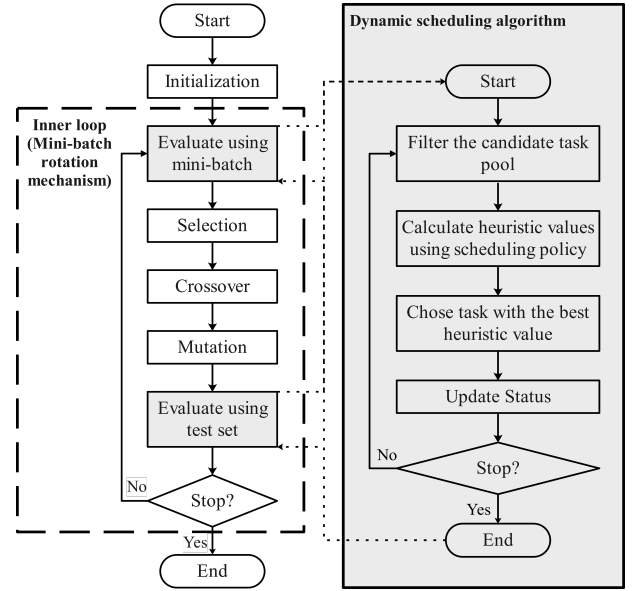

Fig. 1. The flowchart of GPHH designed in this paper

### B. Dynamic scheduling algorithm based on scheduling strategy

The dynamic scheduling algorithm is employed to evaluate scheduling policies, enabling the construction of feasible solutions based on policies in any scheduling scenario. Initially, the solution set is empty, and all to be observed are added to candidate task pool $U$. During the decision-making process, $Filter()$ is designed to timely remove tasks in $U$ that violate constraints. Simultaneously, $BinaryPredecessor()$ is proposed to compute the earliest observable time window for each candidate task. When scheduling begins or completes the previous task, the scheduling policy can be used to calculate the heuristic values of candidate tasks in $U$ and select the optimal one as the next observation task. Since the memory consumption is uncertain, actual consumption may exceed the onboard capacity, resulting in imaging failure. Imaging failure will terminate the scheduling and no benefit will be obtained from the task.

The above dynamic scheduling algorithm has the following advantages:

- $Filter()$ enables rapid updates to $U$ by filtering out unfeasible tasks, ensuring the feasibility of solutions. It enhances decision-making efficiency while reducing noise interference in decision-making.
- $BinaryPredecessor()$ can quickly and accurately calculate the earliest observable time of tasks based on binary search. It ensures that the final scheduling plan meets the timing constraints. It also reserves time window resources for subsequent scheduling, which is conducive to improving resource utilization and task completion rates [19].

To further improve the running efficiency of $Filter()$, a multi-branch hierarchical progressive framework is proposed. Candidate tasks in $U$ are sorted in ascending order according to the start time of the visible time window. The function process of $Filter()$ is as follows:

1) **Pruning:** If the current time $t_{now}$ plus the maximum transition time does not exceed the start time of visible time window for task $i$, this indicates the satellite can definitely complete the attitude transition to observe task $i$. The maximum transition time is determined by the satellite performance [20]. If task $i$ satisfied the inequality, then all subsequent tasks including it will not violate the timing constraints (tasks with imaging durations exceeding the width of the visible time window have been excluded).

2) **Service timeout check:** If $t_{now}$ plus $dur_i$ exceeds the end time of visible time window of task $i$, task $i$ is clearly cannot be completely imaged.

3) **Capacity constraint check:** The memory slack factor $slack_m$ is introduced to adjust dynamic decision preferences. If the expected memory consumption of task $i$ multiplied by the relaxation factor exceeds the remaining onboard memory capacity, task $i$ is excluded. The greater the value of $slack_m$, the higher the likelihood that the task will be excluded.

4) **Earliest observable time check:** A binary precedence scheduling function will be designed to get the earliest observable time for task $i$. If task $i$ has no observable time, it means that observing the task at any time within its visible time window violates either the attitude transition constraint or the time window constraint.

$BinaryPredecessor()$ can provide precise observation time windows. Assuming that the satellite observes task $j$ after completing the observation of task $i$. Define the attitude transition constraint $tc: (oe_i, os_j, Trans)$ and its time delay function $Delay_{tc}()$ as Eq. (18). Regarding the mathematical characterization of attitude transition constraint for AEOS, Pralet and Verfaillie demonstrated that it satisfies time-delay monotonicity [21]. It indicates that $Delay_{tc}()$ monotonically decreases, and the intersection point of the curve with the horizontal axis corresponds to the earliest observable time for the respective task.

$$Delay_{tc}(oe_i, os_j) =$$
$$oe_i + Trans\left(att_{oe_i,i}, att_{os_j,j}\right) - os_j \quad (18)$$

Therefore, we propose a binary precedence scheduling mechanism. Binary search is a well-known method for efficiently locating specific elements on a monotonic function, with a time complexity of O(log $n$). It works by iteratively adjusting the left and right pointers within a monotonic function until convergence. The time corresponding to the coinciding pointer is the earliest observable time. If the time delay of the left pointer is negative, the time corresponding to the left pointer is the earliest observable time. Conversely, if the time delay of the right pointer is positive, it indicates that the task does not have an observable time window.

To ensure that the earliest observable time window simultaneously satisfies both the attitude transition and visibility time window constraints, the search process is divided into two stages. The left endpoint and right endpoint are denoted by $l$ and $r$, respectively. The first stage determines the limit position of $r$ to ensure that any time between $l$ and $r$ can be used as the earliest observable time and still complete imaging within the visible time window. The second stage finds the valid earliest observable time without violating the attitude transition constraint.

*C. Individual representation and initialization*

In GPHH, each scheduling policy is represented as a tree that can be uniquely converted into a mathematical expression. This expression consists of state features and functions. Fig. 2 shows an example of tree representation, and its mathematical expression is $(A + \max(B, C))$. Before making a decision, the heuristic value of each candidate task is calculated using the scheduling policy. Subsequently, the task with the optimal heuristic value is selected to observe.

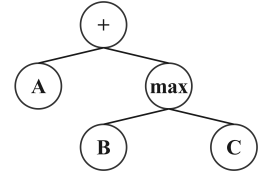

Fig. 2. Tree representation of scheduling policy

Individual initialization in GPHH uses a hybrid method called Half and Half. It combines the Grow method and the Full method, one of which is selected with equal probability during individual generation. The primary difference between the two methods lies in the depth of leaf nodes. In the Grow method, nodes are randomly selected from the function set and terminal set. In the Full method, intermediate nodes are selected exclusively from the function set, resulting in all leaf nodes having the same depth.

*D. Genetic Operators*

The genetic operators designed for the evolutionary stage are as follows:

- **Selection:** The tournament method is employed with a tournament size of $T_{size}$. First, $T_{size}$ individuals are sampled with equal probability and without replacement from the parent population. Next, the best performers are selected to join the offspring population. If there are multiple optimal individuals exist, the winner is chosen randomly from among them.

- **Crossover:** The single-point crossover operator is adopted. First, two individuals are randomly selected from the offspring population. Next, crossover points are randomly chosen from their non-leaf nodes. The subtrees rooted at these crossover points are then exchanged between the two individuals to generate two new individuals. The selected ones are finally replaced by the newly produced individuals.

- **Mutation:** The uniform mutation operator is used. First, each individual undergoes mutation at random. If mutation is required, a node is randomly selected from the tree structure. Then, a new tree is generated using the Half and Half method and replaces the selected node. It is important to note that the minimum and maximum depth settings for the new tree produced by this operator may differ from those specified during individual initialization.

*E. Mini-batch rotation training mechanism*

It is obvious that using all training instances in each round of training phase would be inefficient. To address the issue, we adopt a mini-batch rotation mechanism. It can significantly reduce computational costs and effectively prevent model overfitting [22]. The training set is divided into multiple subsets of equal size, i.e., mini-batches. Only one mini-batch is used per round. To avoid usage imbalances among different mini-batches caused by random sampling, we employ a rotation mechanism [16]. It will cycle through all mini-batches until the algorithm terminates.

## IV. Experimental Study

*A. Experimental Environment*

In this study, instance generation and algorithm testing were conducted in separate environments.

- **Instance generation:** Instance generation was performed on a laptop with the following specifications: CPU: 13th Gen Intel(R) Core(TM) i9-13980HX, clock speed: 2.20 GHz, RAM: 15.6 GB. MATLAB R2020b and the Satellite Toolkit (STK) 12 were used to simulate satellite data and generate the instance set.

- **Algorithm testing:** PyCharm was used to configure an SSH remote server on the cluster for algorithm testing. The cluster uses AMD Opteron 4184 processors (2.8 GHz, 2 GB RAM) running Debian 11. The algorithm programming language is Python, version 3.11.

*B. Instance Generation*

UAEOSSP requires multiple instances to represent the same scheduling scenario under different environments. As no benchmark exists for this novel problem, new instances must be generated to facilitate algorithm testing. Notably, static AEOSSP instances can serve as a basis for deriving UAEOSSP instances.

First, a static AEOSSP instance set must be generated. This paper builds a simulation scheduling scenario based on STK. The experimental scenario time window is set from 2024-09-01 00:00:00 to 2024-09-01 24:00:00 (UTC). The design references for the AEOS and task-related parameters are based on existing research. The maximum pitch angle and roll angle maneuvering range of the satellite are set to [-27,27] (deg) [23]. The expected rate for imaging write code is set to 3.5 (GB/s) [24]. The relevant parameters involved in the attitude transition time function *Trans()* are shown in the TABLE II [15]. The orbital parameters are set using the classic six-parameter orbital equation, with the six orbital parameters being the semi-major axis ($a$), eccentricity ($e$), inclination ($i$), perigee argument ($\omega$), ascending node longitude (RAAN), and true anomaly ($m$). Based on relevant literature, the satellite's orbital altitude is set to 500 km, the orbital inclination to 100 degrees, and the argument of perigee to 0 degrees[1], [15] . The satellite orbital parameters are listed in the TABLE III.

TABLE II. ATTITUDE TRANSITION TIME FUNCTION PARAMETERS

| Segment $i$ | $a_i$ | $v_i$ | $\theta_{i0}$ | $\theta_{i1}$ |
|---|---|---|---|---|
| Segment 1 | 5 | 1 | 0 | 15 |
| Segment 2 | 10 | 2 | 15 | 40 |
| Segment 3 | 16 | 2.5 | 40 | 90 |
| Segment 4 | 22 | 3 | 90 | - |

TABLE III. SATELLITE ORBITAL PARAMETERS

| Satellite | $a$ (m) | $e$ (deg) | $i$ (deg) |
|---|---|---|---|
| Satellite 1 | 6878137 | 0.00 | 100.00 |
| **Satellite** | $\omega$ (deg) | RAAN (deg) | $m$ (deg) |
| Satellite 1 | 0.00 | 360.00 | 360.00 |

To generate diverse scheduling scenarios, this paper uses different combinations of scenario parameters to create an AEOSSP instance set with distinct characteristics. These parameters include the number of tasks ($NT$), the scheduling stop time ($ST$), and the maximum memory capacity ($MMC$). $NT$ determines the scale of the tasks in the scenario directly and is set to 50, 100, 150, or 200. $ST$ affects the distribution density of the tasks and is set to 2000, 4000, and 6000. $MMC$ affects the capacity of onboard memory. Because different task sizes have varying memory requirements, the $NT$ and $MMC$ settings is designed as follows: When $NT$ is set to 50, the $MMC$ value set is {1024, 2048, 4096}. When $NT$ is set to 100 or 150, the $MMC$ value set is {2048, 4096, 6144}. When $NT$ is set to 200, the $MMC$ value set is {4096, 6144, 8192}.

The task parameters refer to the experimental design of Chu et al. on AEOS scheduling [25]. UAEOSSP considers uncertainties in task profit, resource consumption, and task visibility. The actual task profit and resource consumption are generated by referring to the research of Mei et al. [26]. Mei et al. employed a Gamma random distribution with a shape parameter of 20 when considering the actual requirements of tasks with an expected value of 20. The Gamma distribution is commonly used to generate non-negative random variables. It can be represented as $Gm(\alpha, \beta)$, where $\alpha$ is the shape parameter and $\beta$ is the scale parameter. The mean of $Gm(\alpha, \beta)$ is $\alpha\beta$, and the variance is $\alpha\beta^2$. Given a known expected value, the distribution characteristics of $Gm(\alpha, \beta)$ can be altered by adjusting the parameters. This experiment uses a Gamma distribution with $\alpha_{profit} = 30$ to randomly generate the actual task profit. Since the actual write code rate does not change significantly, a Gamma distribution with $\alpha_{cr} = 350$ is used to generate it. Cloud coverage only considers two cases: no coverage and full coverage [5]. The observable probability for each task is $prob_{cloud}$. TABLE IV shows the parameters for generating tasks and uncertain variables.

TABLE IV. PARAMETERS FOR GENERATING TASKS AND UNCERTAIN VARIABLES

| Task | | Uncertainty | |
|---|---|---|---|
| Task Distribution | Random uniform | $\alpha_{profit}$ | 30 |
| Imaging time | $N(25,9)$ | $\alpha_{cr}$ | 350 |
| Yield distribution | $N(2 \times dur_i, 100)$ | $prob_{cloud}$ | {0.1,0.2,0.3} |

*C. Parameter Setting*

The nodes of GP trees consist of elements from both the terminal set and the function set. The design of these elements has a direct impact on the performance of the algorithm and the effectiveness of the generation policies. The terminal set provides the GP trees with input, while the function set provides them with computational logic.

The terminal set includes three types of key features: profit, memory and time. All elements are normalized to [0,1] to avoid dimensional inconsistencies. TABLE V presents the design of the terminal set for this study.

TABLE V. TERMINAL SET DESIGN

| Related | Notation | Description |
|---|---|---|
| Profit | RP | Real profit. This feature is obtained through the Min-Max method [27]. |
| | RPPU | Real profit per unit of time. This feature is obtained through the Min-Max method. |
| Memory | EMC | Expected memory consumption. This feature is obtained through the Min-Max method. |
| | EMO | Expected memory occupancy. This feature is obtained by dividing the expected memory consumption by the remaining onboard memory. |
| | RMP | Remaining memory percentage. This feature is obtained by dividing the current onboard memory remaining amount by the maximum memory capacity. |
| Time | CT | Current time. This feature is obtained by dividing the current time by the total scheduling time. |
| | TIST | Task imaging start time. This feature is calculated as Eq. (19). $t_{now}$ is the current time; $T$ is the total scheduling time; $os_i$ is the earliest observable time of the task; $c$ is a small constant. |
| | RTP | Remaining task percentage. This feature is obtained by dividing the number of current candidate tasks by the total number of tasks in the current scheduling scenario. |
| | FR | Full ranking. All tasks are sorted in order of their start times within the visible time window, from earliest to latest. This feature is obtained by dividing the position of the task in the entire task list by the total number of tasks. |
| | RR | Relative Ranking. Unlike FR, this feature considers the order of tasks within the candidate task pool $U$. The feature is calculated by dividing the order by the number of candidate tasks in $U$. |
| Other | C | Constant. This constant is generated from a uniform random distribution in the range [-1, 1]. |

$$TIS_i = \frac{os_i - t_{now} + c}{T - t_{now} + c} \quad (19)$$

According to [10], [16] and preliminary experimental studies, the function set is defined as $\{+, -, \times, \div, \max, \min, \sin\}$. It is imperative to note that all functions are safeguarded to prevent errors. Specifically, when the result of an operation is infinite ($\infty$), the function returns 1.

In the course of the evolutionary process, we use a training mechanism involving mini-batch rotation. The 100 training instances were partitioned into 20 mini-batches, with each mini-batch comprising 5 instances. Furthermore, the GP parameter settings are outlined in TABLE VI.

For the small-batch rotation training mechanism, the 100 training instances are divided into 20 small batches, so each small batch contains 5 instances. Other parameter settings of the GP are listed in TABLE VI.

TABLE VI. GP PARAMETERS

| Parameter | Value | Parameter | Value |
|---|---|---|---|
| Population size | 20 | Tournament size | 4 |
| Number of iterations | 60 | Maximum depth of the tree | 1 |
| Crossover probability | 0.85 | Initial maximum depth of generated tree | 6 |
| Mutation probability | 0.15 | Maximum depth of mutated generated tree | 4 |

*D. Comparative Experimental Analysis*

To demonstrate the efficacy of the proposed GPHH, we will compare it with Look-Ahead Heuristics (LAHs) and Manually Designed Heuristics (MDHs). Both methods are heuristic approaches designed based on expert knowledge. The following provides an introduction to LAHs and MDHs.

LAHs construct solutions by looking ahead to future tasks. During dynamic scheduling, they select tasks based on various greedy rules [20]. To better address the UAEOSSP, this paper modifies LAHS in three aspects.

- **Look-ahead step size.** The look-ahead step size is a positive integer. Depending on its value, different numbers of tasks can be looked ahead. Only tasks within the look-ahead window will be selected until scheduling ends.

- **Filtering mechanism.** Before making a decision, the tasks that violate the constraints must be filtered out to ensure the feasibility of the solution.

- **Decision module.** The decision module represents the rule for selection. Different decision modules will result in varying scheduling preferences. Here, we will design three decision modules for the LAHs.

Based on different decision modules, three types of LAHs are presented as shown in TABLE VII. Experiments revealed that different look-ahead step sizes significantly impact the dynamic scheduling performance of AEOS. For most UAEOSSP instance sets, the expected total profit exhibits a pattern of initially increasing and then declining as the look-ahead step size increases. LAH2 and LAH3 employ different look-ahead step sizes, and the optimal result is taken as the performance of this method.

TABLE VII. LOOK-AHEAD HEURISTICS

| Algorithm | Description |
|---|---|
| LAH1 | The look-ahead step size of LAH1 is 1, making it a greedy algorithm that observes the most nearest observable task [28]. |
| LAH2 | LAH2 selects the task with the highest actual profit among look-ahead tasks. The look-ahead step size ranges from 2 to 20 (when set to 1, it corresponds to LAH1). |
| LAH3 | LAH3 selects the task with the highest value of profit divided by imaging time among look-ahead tasks. The look-ahead step size ranges from 2 to 20. |

Manually Designed Heuristics (MDHs) are methods for designing rules based on expert knowledge. Compared to ODS generated using GPHH, MDHs perform less effectively but generally offer greater interpretability. This experiment will design MDHs that can be directly adapted to the dynamic scheduling algorithm, as shown in TABLE VIII. The previous observed task is $pre$, and the current attitude information for $t_{pre}$ is $att_{pre}$.

TABLE VIII. MANUALLY DESIGNED HEURISTICS

| Algorithm | Introduction |
|---|---|
| MDH1 | The value density strategy is inspired by one of the classic strategies for the 0-1 knapsack problem [29]. The higher the value density, the higher the priority. The value density is calculated as Eq. (20). |
| MDH2 | This strategy is inspired by a greedy algorithm used to construct an initial solution for the TSP, where the nearest node is selected as the next travel destination [30]. In UAEOSSP, the task with the shortest required attitude transition time is prioritized. The attitude transition time is calculated as Eq. (21). |
| MDH3 | When the remaining memory is less than half, MDH1 is adopted; otherwise, MDH2 is adopted. Define the current remaining memory as $mmc_{now}$. The selection rule for this algorithm is shown as Eq. (22). |

$$MDHS_1(i) = \frac{\bar{p}_i(env)}{dur_i + Trans(att_{pre}, att_{os_i,i})} \quad (20)$$

$$MDHS_2(i) = \max\{Trans(att_{pre}, att_{os_i,i}), os_i - t_{pre}\} \quad (21)$$

$$MDHS_3(i) = \begin{cases} MDHS_1(i), mmc_{now} < \text{MMC}/2 \\ MDHS_2(i), mmc_{now} \geq \text{MMC}/2 \end{cases} \quad (22)$$

The performance of GPHH and the comparison algorithms is evaluated on 108 UAEOSSP instance sets. Execute GPHH using varied random number seeds and document the outcomes. The scheduling policy that achieves the highest performance on the test set is selected as the final outcome of GPHH. TABLE IX displays the aggregate of the best overall gains for LAHs, MDHs, and GPHH across various task scales.

TABLE IX. PERFORMANCE OF THE THREE ALGORITHMS

| Task scale | LAHs | MDHs | GPHH |
|---|---|---|---|
| 50 | 35122.52 | 34025.12 | 36604.74 |
| 100 | 59248.89 | 59214.28 | 63338.07 |
| 150 | 64303.45 | 62784.45 | 68687.13 |
| 200 | 89156.65 | 91176.16 | 94965.36 |

The outcomes reveal that the total optimal profit of GPHH markedly surpasses that of the comparison algorithms. Furthermore, $Gap(A, B)$ is defined as the gap between A and B, and it is calculated as Eq. (23). Fig. 3 displays the $Gap$ between GPHH and the comparison algorithms on each instance set of UAEOSSP. It is obvious that the value of $Gap(\text{GPHH},*)$ being greater than 0 indicates that the GPHH is superior. The greater the value of $Gap(GPHH,*)$, the more outstanding the relative performance of GPHH.

$$Gap(A, B) = \frac{Output_A - Output_B}{Output_B} \quad (23)$$

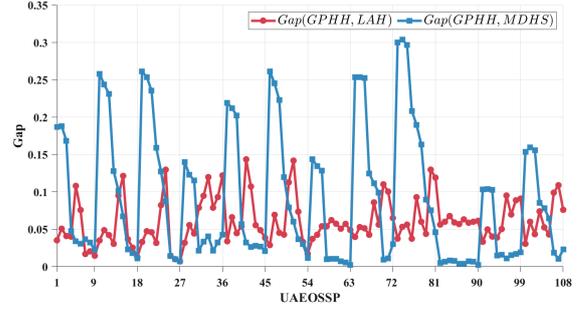

Fig. 3. The gap between the GPHH and LAH, MDHS in optimal total profit

The outcomes in Fig. 3 indicate that GPHH achieves optimal performance across all UAEOSSP instance sets. The gap between the optimal, average, and worst performance obtained from multiple runs of GPHH and the optimal performance of LAHs corresponds to 6.05%, 5.03%, and 3.63%, respectively. The corresponding values for GPHH and MDHs are 9.20%, 8.14%, and 6.67%. To intuitively compare the decision-making performance of evolved scheduling policies and manually designed heuristics, test instances are randomly sampled from scenarios with different task scales.

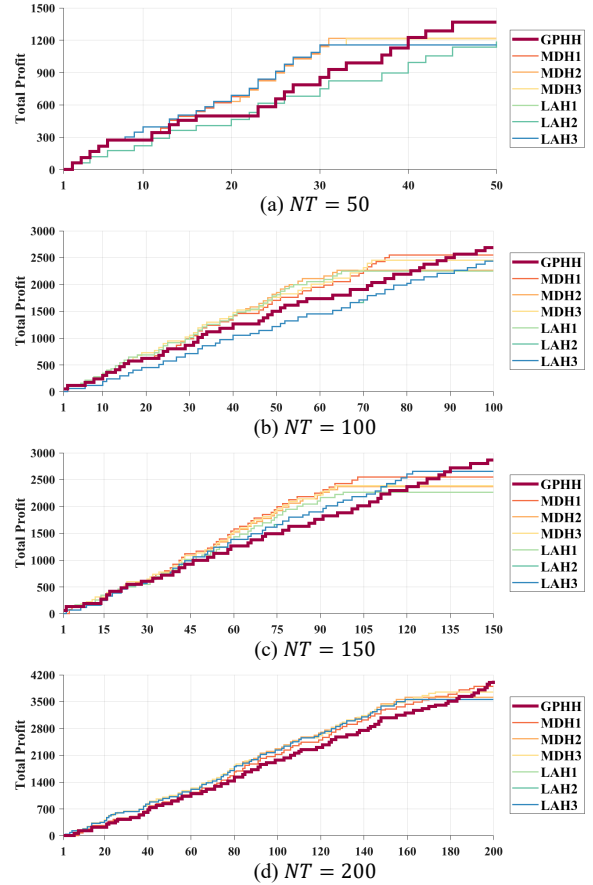

Fig. 4. Ladder diagram of the total profit in decision processes for different methods on test instances

The bold red curve in Fig. 4 represents the performance of GPHH. Throughout the entire scheduling process, this curve exhibits more consistent variation compared to other curves and surpasses all others by the end of scheduling horizon. It demonstrates that the scheduling policy evolved by GPHH possesses stronger global optimization capability than the alternative approaches.

CONCLUSION

In satellite scheduling, environmental factors such as cloud cover, memory consumption, and task profit are inherently uncertain. This makes imaging satellite scheduling a critical yet underexplored practical issue. To better align with reality, we propose a novel problem called the Uncertain Agile Earth Observation Satellite Scheduling Problem (UAEOSSP). It considers various uncertainties including task profit, memory consumption and visibility. A stochastic programming model is built for UAEOSSP. To obtain robust solutions with the highest expected total profit, we propose an efficient Genetic Programming Hyper-Heuristic (GPHH). This method is utilized to automatically evolve an excellent scheduling policy, which can be embedded into an online dynamic construction algorithm to generate solutions. Experimental results demonstrate the superior effectiveness of our method. The scheduling policies generated by GPHH outperforms advanced the well-designed LAHs and MDHs.

However, experiments also reveal that the evolved scheduling policies generally exhibit significant shortcomings at scale. The complex structures engender a reduction in interpretability, making it harder for operators to understand and trust. Consequently, future research will concentrate on designing improved methods to generate highly interpretable policies by enhancing the simplicity of reduction strategies and refining key scheduling logic. We will also request actual satellite scheduling data from relevant authorities to validate the practicality of our algorithm.